\documentclass[letterpaper, 10 pt, conference]{ieeeconf}  

\IEEEoverridecommandlockouts                              

\usepackage{graphicx}
\usepackage{amsmath}
\usepackage{amsfonts}
\usepackage{bm}
\usepackage{booktabs}

\newcommand{\secref}{Section~\ref}
\newcommand{\figref}{Figure~\ref}
\newcommand{\tabref}{Table~\ref}

\title{\LARGE \bf
Plan-and-Act using Large Language Models for Interactive Agreement
}

\author{Kazuhiro Sasabuchi$^{1}$, Naoki Wake$^{1}$, Atsushi Kanehira$^{1}$, Jun Takamatsu$^{1}$, and Katsushi Ikeuchi$^{1}$
\thanks{*This work was not supported by any organization}
\thanks{$^{1}$All authors are with Applied Robotics Research, Microsoft, Redmond, WA, USA
        {\tt\small Kazuhiro.Sasabuchi@microsoft.com}}%
}

\begin{document}

\maketitle
\thispagestyle{empty}
\pagestyle{empty}

\begin{abstract}

Recent large language models (LLMs) are capable of planning robot actions. In this paper, we explore how LLMs can be used for planning actions with tasks involving situational human-robot interaction (HRI). A key problem of applying LLMs in situational HRI is balancing between ``respecting the current human's activity" and ``prioritizing the robot's task," as well as understanding the timing of when to use the LLM to generate an action plan. In this paper, we propose a necessary plan-and-act skill design to solve the above problems. We show that a critical factor for enabling a robot to switch between passive / active interaction behavior is to provide the LLM with an action text about the current robot's action. We also show that a second-stage question to the LLM (about the next timing to call the LLM) is necessary for planning actions at an appropriate timing. The skill design is applied to an Engage skill and is tested on four distinct interaction scenarios. We show that by using the skill design, LLMs can be leveraged to easily scale to different HRI scenarios with a reasonable success rate reaching 90\% on the test scenarios.

\end{abstract}

\section{INTRODUCTION}

\begin{figure}[t]
\centering
\includegraphics[width=1.0\columnwidth]{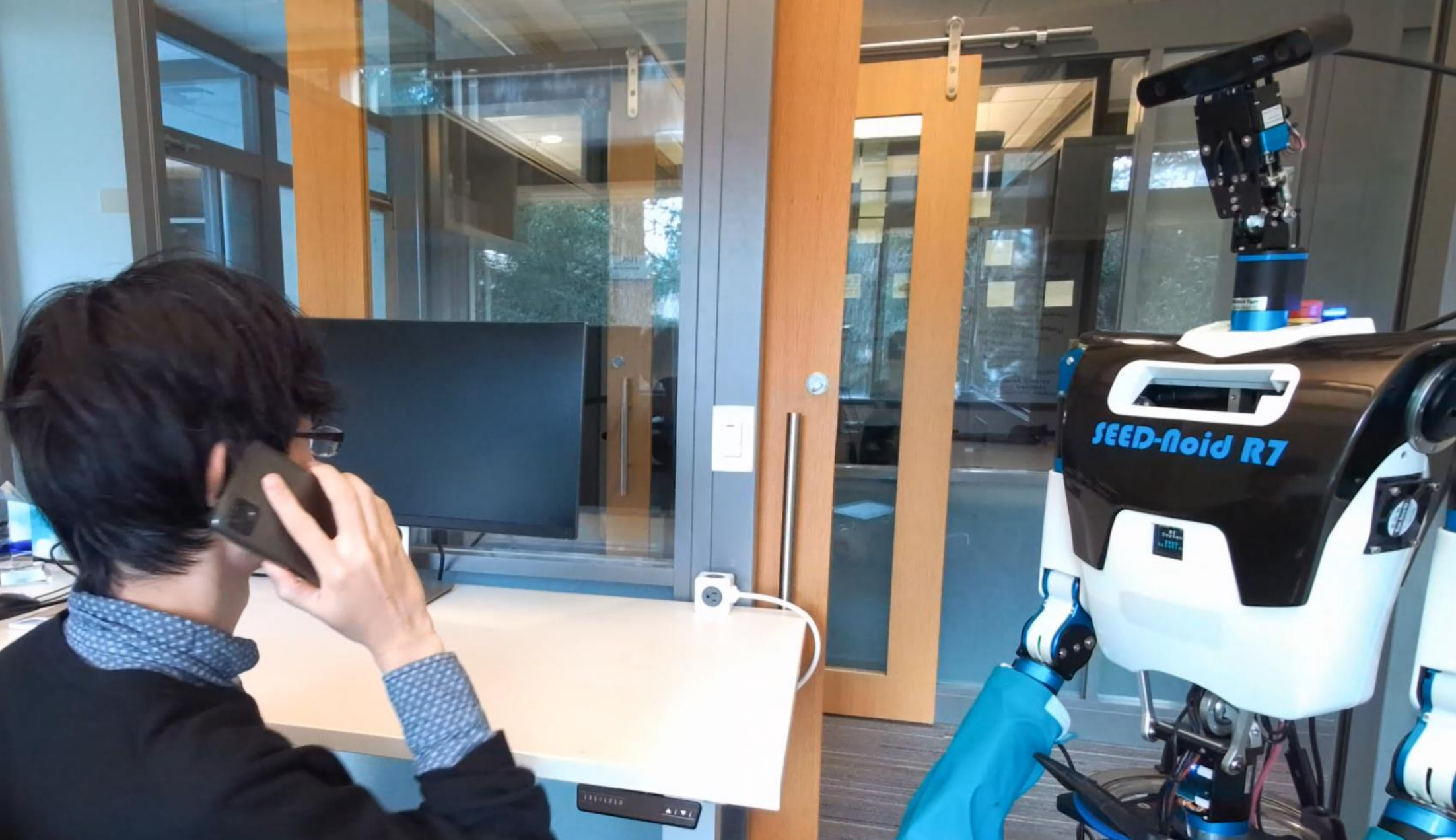}
\caption{A situation where a robot is trying to execute a ``speak" but should re-plan its actions based on the runtime HRI situation.}
\label{fig:intro}
\vspace{-3mm}
\end{figure}

Recent Large Language Models (LLMs) are capable of reasoning and providing answers to text-based questions \cite{binz2023using}. The technology has been applied in robotics for generating task plans and our previous work \cite{wake2024gpt} has focused on applying LLMs for offline planning (plans generated prior to running the robot's task). In this paper, we explore how to apply LLMs in robotic systems that involve runtime planning (plans generated as the robot executes its task), which is often the case for tasks involving human-robot interaction (HRI). In a HRI situation, whether a robot can execute a task is dependent on the situation of the human. For example, in a speak task happening at an engaging stage (\figref{fig:intro}), the human might be under a phone call and not ready to engage nor ready to listen to the robot's speech. In such situations, the robot must re-plan its actions to wait until the person finishes the phone call. The job of the LLM in this case is to generate actions \textit{``wait"} then \textit{``speak"} based on the observation of the human situation \textit{``under a phone call"} and the robot's goal \textit{``perform a speak task."}

Similar to offline planning, LLMs should be able to provide runtime plans. 
However, the two main differences between offline planning and runtime planning for an HRI situation are: (1) the need for taking into account the human situation which may not always be aligned with the robot's current goal (the agreeing to interact problem \cite{sasabuchi2018agreeing}), (2) planning happens as events occur during runtime and requires figuring out the timing of when to generate a new plan. 

Regarding the first problem, our investigation \cite{sasabuchi2025agreeing} has shown that recent LLMs like GPT-4o (2024-08-06 model), with its large data of common sense, are indeed capable of generating robot actions while incorporating both the human situation and the robot's goal. However, we have also found that generating a reasonable action requires providing the right observations (description of the human's activity and gaze directions) and that LLMs have difficulty balancing between \textit{``respecting the current human's activity"} and \textit{``prioritizing the robot's goal/task."}

In this paper, we introduce a necessary skill design to address the above challenges and to achieve runtime plan-and-act in situational HRI.
The key factors of the design are three-folds. First, the skill uses a bottom-up action set which enables flexible integration with outputs generated from the LLM. Second, the skill uses an event manager which asks the LLM a second-stage question on the returned actions to figure out the timing to generate the next plan. Third, the skill passes action texts about the current robot's action to the LLM, which can facilitate the robot to prioritize its task and drastically improve the action outputs.

For evaluation of the skill design, we will use GPT-4o (2024-08-06 model), which is one of the publicly accessible models with one of the highest performance in the area \cite{islam2024gpt, abdin2024phi}. While some of the results are limited to the tested model, the findings are somewhat general, and the design should apply to other off-the-shelf language models.

\section{Related Works}
\label{related_works}

The problem being solved in the runtime planning is to decide a robot action based on the situation of the human. Such planning occurs especially when there is a misalignment between the robot's goal and the human situation \cite{sasabuchi2018agreeing}. The problem is relevant to the problem of understanding human signs such as ``engagement," which refers to the interaction interest of a person toward a machine \cite{sidner2005explorations, bohus2014directions, anzalone2015evaluating}, human interruptibility \cite{banerjee2017temporal}, and non-engagement \cite{rossi2018disappearing}. However, the primary focus of the listed research is on detecting human signs. 
By leveraging LLMs, both the understanding of human signs and automated planning of actions can be solved at once using a single pre-trained model.


Several recent works have focused on using pre-trained LLMs in HRI. \cite{kim2024understanding} has investigated how LLM-powered robots compare with text-based agents and voice agents. \cite{wang2024lami} has used LLM to plan human-assistive actions from multi-modal scene perception. \cite{lee2023developing} has used LLM to generate non-verbal cues in a conversation system. \cite{verma2024theory} has studied mental-model reasoning abilities of LLMs to improve behavior synthesis in HRI. Our work \cite{sasabuchi2025agreeing} has investigated LLM's capability to solve the interaction agreement problem; this paper adds on top of the work to leverage such capabilities and apply as a skill that can be executed on a robot system.

Of the LLM research, the closest work is \cite{wang2024lami} which implements an HRI system leveraging LLMs and producing robot actions for proactive assistance. While there are some similarities, the focus of proactive assistance is aligning the robot's task to the human's situation. This work includes some reverse conditions where the robot plans to act so that it can enable a human to align with the robot's task (such as by catching attention). The difference in the problem leads to different designs such as balancing between ``respecting the human's activity" and ``prioritizing the robot's goal/task," incorporating initial running robot actions, context about the robot's action, and generating observe timings associated with the actions.


\section{Method}
\label{method}

\begin{figure*}[t]
\centering
\includegraphics[width=1.8\columnwidth]{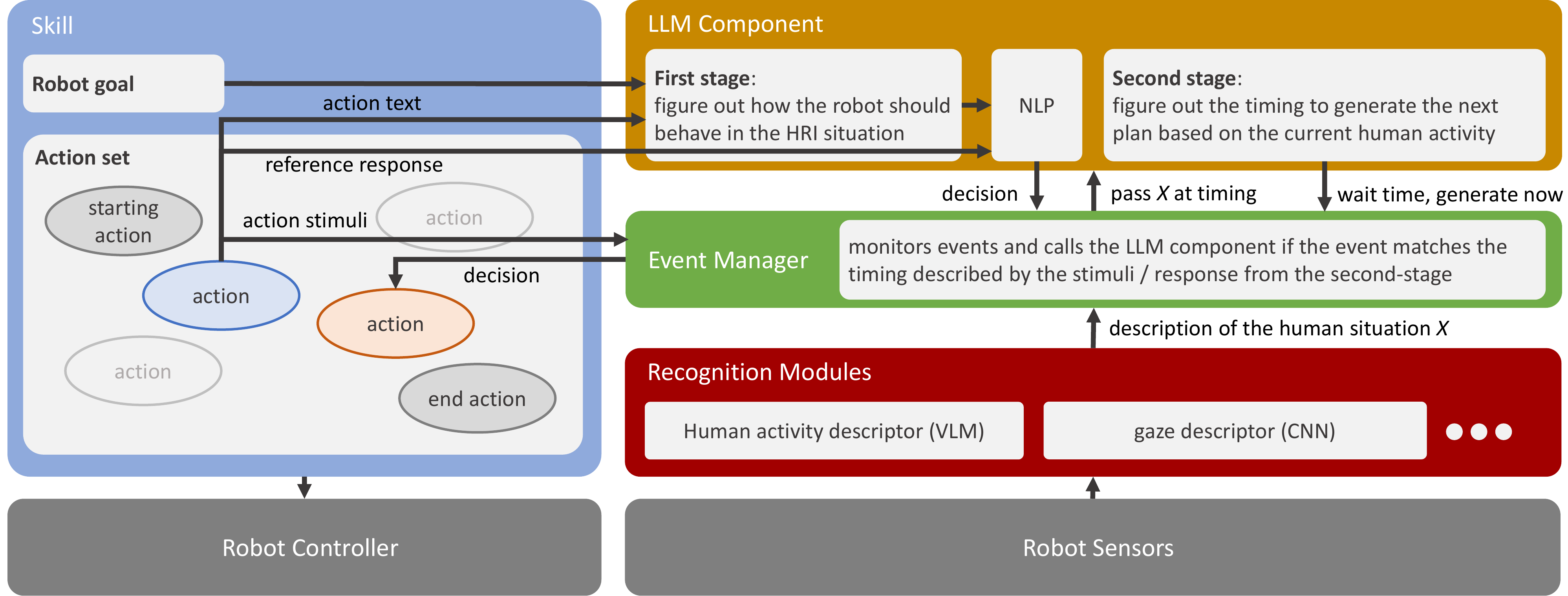}
\caption{Proposed skill design. The action in blue indicates the current action. The action in orange indicates the next action decided using the LLM and event manager. The skill begins from the starting action and changes the action based on the LLM's response until reaching the end action. Timing to call the LLM is managed by the event manager, which uses situational changes from the recognition modules and second-stage questions to decide the timing.}
\label{fig:method}
\vspace{-3mm}
\end{figure*}

In this work, we leverage LLMs to generate an appropriate action plan for the robot to accomplish a task involving an HRI situation. The plan by the LLM is generated based on the robot's task goal and observed human situation at runtime. In order to leverage the LLM, we design the plan-and-act capability in a form of a skill \cite{wake2021learning, ikeuchi2024semantic, takamatsu2024designing}. A skill has a clear start and goal state which allows combining with other skills to form a larger operation \cite{wake2023chatgpt}. Unlike manipulation skills however, the actions to achieve the goal state from the start state within a skill 
is dependent on the current situation of the human (thus, we solve the action decision by calling the LLM on runtime). An overview of the proposed skill design is shown in \figref{fig:method}. Below, we will explain each component of the design in detail.

\subsection{Skill Involving HRI}
\label{skill_design}

A robot's skill is used to achieve a defined goal state from some start state. The start state is usually the end state achieved by a previous skill (e.g., a robot at a specific location such as near a person after running a navigation skill). A skill in its most simplest form can run a single action that enables the robot to achieve the goal state. We will call this action the \textbf{end action}. However, in an HRI-involved situation, whether an end action can be executed is dependent on the situation of the human. Therefore, we add an additional action which allows the robot to observe the situation first, which we call the \textbf{starting action}.

If the situation of the human is aligned with the robot's goal, the robot can directly execute the end action after the starting action. However, if the situation is not aligned, the robot will require some additional actions before executing the end action. To run these actions from the skill, we assume a preset of actions, which we call the \textbf{action set}, is implemented using commands that can be sent to the robot. Which action from the action set to use (including whether to execute the end action) is solved by the LLM component explained later. 

In addition, the skill has a defined field \textbf{robot goal} which is used by the LLM component to generate the action plan. The robot goal should be some text description relevant to the goal state of the skill and may also include some extra context such as the environment the skill is being executed.

\subsection{LLM Component}
\label{llm_component}

The LLM component decides the action the robot should execute based on the observed situation about the human and the robot's goal. To obtain the situation of the human, external recognition modules are used to generate a text description about an observed scene from the robot's camera. In this paper, we specifically concatenate the outputs of the following modules: (1) a human activity descriptor module, which can be implemented using a vision language model (VLM) that runs on an edge machine \cite{abdin2024phi}, (2) a gaze descriptor (description about whether the human is looking at the robot or not), which can be implemented using a pretrained convolution neural network (CNN) \cite{6drepnet2024}. The choice of using these modules are based on results from our preliminary experiment \cite{sasabuchi2025agreeing}.

In addition to the robot's goal defined by the skill, we add text about the current robot's action which we call the \textbf{action text}. Although the LLM component itself can work without the action text, we show in \secref{experiment} that adding such text drastically improves the LLM's performance. We will refer to the concatenated text including both the robot's goal and the action text as the \textbf{robot's situation}.

\begin{figure}[t]
\centering
\includegraphics[width=1.0\columnwidth]{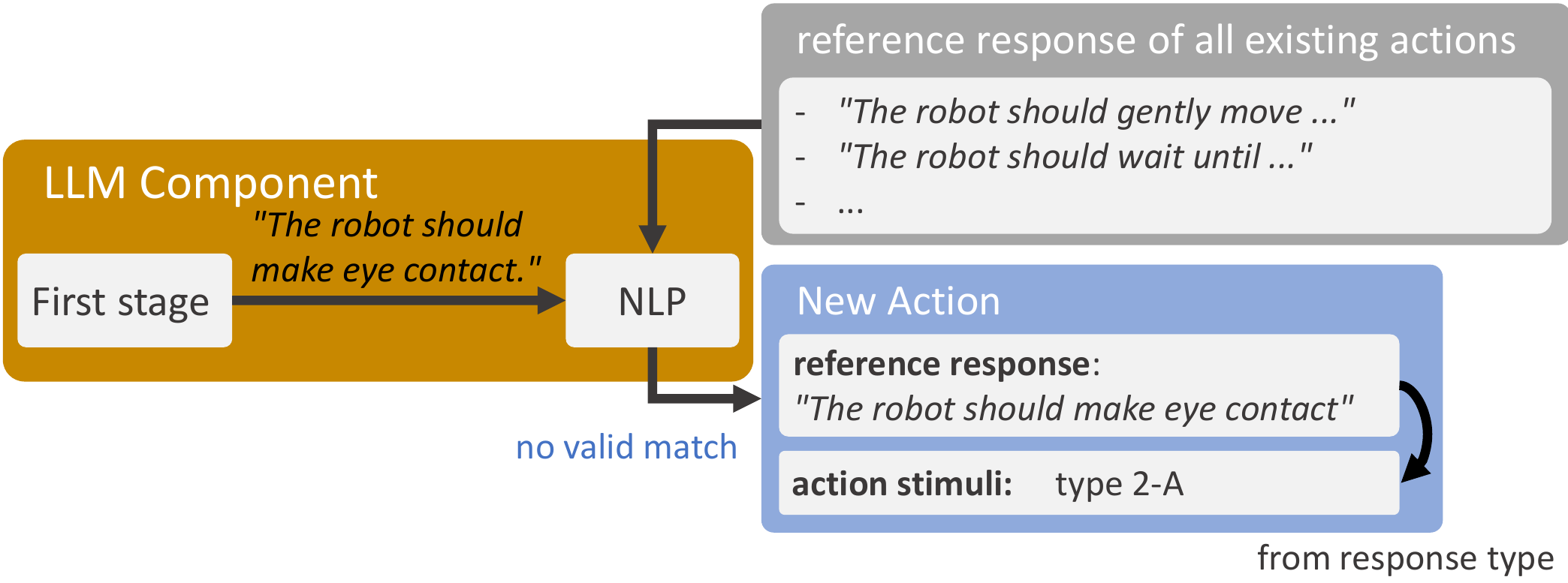}
\caption{Bottom-up approach for registering new actions in the action set. The figure shows an example of creating a new action "eye contact" based on the response from the LLM, existing actions, and the NLP component. }
\label{fig:bottom_up}
\vspace{-5mm}
\end{figure}

Using the description about the human situation and the robot's situation, the LLM outputs a plan about the actions the robot should take using the prompt shown in \tabref{tab:llm_policy_prompt}, where \textit{U} is the robot's situation and \textit{X} is the concatenated text about the human's situation. History is not preserved as long histories can take a longer time for the LLM to respond and is too slow for using in a system involving HRI (which requires a less than a second response time). In addition, we do not bound the returned actions and instead map the generated response from the LLM to one of the actions from the action set using a natural language processing (NLP) tool. This approach enables a bottom-up approach where the action sets can be designed based on the responses from the LLM. The set can later be expanded if the response for a certain situation does not match any of the existing actions. This response is then saved and used as the \textbf{reference response} of the newly expanded action when performing the NLP mapping in a future execution (\figref{fig:bottom_up}).

\begin{table}[t]
\caption{Prompt used by the LLM component.}
\vspace{-7mm}
\label{tab:llm_policy_prompt}
\begin{center}
\begin{tabular}{|l|p{0.8\linewidth}|}
\hline
user & I will provide a sentence about a human-humanoid interaction situation. Your job is to figure out how the humanoid robot should behave in this situation by returning a dictionary with the keys ``action" and ``reason". Please keep in mind that some people may feel uncomfortable interacting with the robot. If the robot requires some verbal action, also come up with an appropriate phrase.

Below is an example:

- My input: ``The robot is waiting to help a person. The person is approaching the robot. The person is looking at the robot."

- Your output: \{``action": ``The robot should turn towards the person and make eye contact.", ``reason": ``The robot should acknowledge the person's presence. The robot should not speak yet as the person may feel uncomfortable and not engaged yet."\}

Understood? \\
\hline
assistant & Understood! Please provide the sentence about the humanoid-robot interaction situation, and I'll determine the appropriate behavior for the humanoid robot. \\
\hline
user & \textit{U}. \textit{X}. \\
\hline
assistant & \{``action": \textit{A}, ``reason": \textit{Y} \} \\
\hline
\end{tabular}
\end{center}
\vspace{-4mm}
\end{table}

\begin{table*}[t]
\caption{Details of the actions in the engage skill example. Stimuli type refers to the types explained in \secref{stimuli}.}
\vspace{-4mm}
\label{tab:engage_actions}
\begin{center}
\begin{tabular}{|l|p{0.15\linewidth}|p{0.08\linewidth}|p{0.37\linewidth}|p{0.2\linewidth}|}
\hline
action & action text & stimuli & description of implementation & reference response for NLP  \\
\hline
approach & The robot is approaching the person. & type 2-B & Using a cached location of the person, move the robot's position from its current position to 1.5 [m] away from the cached location. A navigation planner is used for moving the robot's position. The action completes once reaching the position but may stop at an earlier timing upon receiving new plans from the LLM. & The robot should gently move into the person's line of sight. \\
\hline
wait & The robot is waiting for the human to finish "$X_a$". & type 1 activity & No commands are sent to the robot. The action finishes upon receiving a new plan from the LLM. $X_a$ is the output of the activity descriptor module at the time when the action begins. & The robot should wait until the person finishes the activity. \\
\hline
wait for cue & The robot is waiting for an eye contact from the person. & type 1 gaze & No commands are sent to the robot. The action finishes upon receiving a new plan from the LLM. & The robot should wait until the person makes eye contact or pauses. \\
\hline
eye contact & The robot made eye contact. & type 2-A  & Using a cached location of the person's face, point the robot's head toward the person. A look-at inverse kinematics solver is used to control the robot's head. The action completes once finished sending the calculated head joint angles. & The robot should make eye contact. \\
\hline
pause & The robot paused its approach. & type 2-A  & Sends a stop request to the navigation planner, finishes upon stop is confirmed. & The robot should briefly pause its task. \\
\hline
speak & N/A (end) & N/A (end)  & Runs a text-to-speech module and outputs the generated speech from the robot's speaker. Speech content is extracted from the LLM's response (speaks a default phrase if the LLM response does not suggest any speech content). & The robot should gently announce its presence and ask if the person would like assistance. \\
\hline
\end{tabular}
\end{center}
\vspace{-4mm}
\end{table*}

\subsection{Event Manager, Timings, and Stimuli}
\label{stimuli}

Unlike offline planning (plans generated prior to running the robot's task) where the LLM is called before an execution, to use LLMs for runtime planning (plans generated as the robot executes its task), one must define the timing to obtain the human situation and generate the action plan (i.e., the timing to call the LLM component).

Based on our preliminary investigation, the type of responses returned by the LLM can be categorized into one of the following: (1) a response containing an action with a specific timing to observe a situation, (2-A) a response containing only the action and the action is a low-level action, (2-B) a response containing only the action and the action is a high-level action.

For category (1), the timing can be defined based on the response from the LLM. For example, with a response \textit{``wait until the person finishes the phone call,"} the LLM can be called once the human activity detector detects a new activity. One issue, however, is that relying on the timing described by the response may cause an infinite wait. For example, in the response \textit{``wait for an eye contact,"} the person may not be aware of the robot and never provide an eye contact. To avoid such cases, we ask the following second-stage question to the LLM to obtain a secondary end timing for the wait: \textit{For the human activity '$X_a$', how long should you wait before catching the person's attention? Please return in the form of a dictionary with keys ``wait\_time" and ``reason". ``wait\_time" should be a float in seconds.}

For category (2), a timing is not included in the response. Instead, the generated actions have a concrete end timing. For example, a response action \textit{``make an eye contact"} ends once the robot moves its head (or eye) toward the person. Once an action ends, the next action must be planned, thus, the LLM component is called at the timing of when the action finishes. However, while this timing is suitable for (2-A) where the low-level actions are short, the high-level actions in (2-B) such as \textit{``slowly approach the person"} can take time to complete. There are two cases where the LLM component should generate a new plan without waiting for an action from (2-B) to be complete. One is when a situational change occurs (e.g., change in the detected activity or change in gaze of looking toward/away from the robot). Another is when there is a chance of losing the person if a plan is not generated soon. The first case is handled by changes in detection. The second case is handled using the following second-stage question: \textit{For the human activity '$X_a$', is there a high chance that the robot will lose the person if did not capture images for $T$ seconds? Please return in the form of a dictionary with keys ``answer" and ``reason". ``answer" should be yes or no.}

The above timings are defined for each action at the timing of saving the reference response (based on the response type of the response). We will refer to this defined timing as the \textbf{action stimuli}. Once a new action begins, the action stimuli is passed to an event manager, which monitors events (such as a change in human activity, elapsed time, etc.), handles secondary questions, and triggers the LLM component if the event matches the timing described by the stimuli (see \figref{fig:method}). Once the output from the LLM component is obtained, the skill switches to the next action based on the generated output.

\subsection{Engage Skill Example}

In this section, we provide an example implementation using the above skill design. The example skill is the Engage skill, which is used to initiate a conversation from an ambiguous situation where the human may or may not be ready to engage with the robot. The skill is used from a starting state that is near but slightly away from the person. The skill is meant to be used as part of a longer operation that continues to a conversation skill (which could also be designed using the proposed skill design to handle situational turn-taking).

The skill contains the following action set based on the responses from the LLM and by using the bottom-up design approach from \secref{llm_component}: \textit{``approach,"} \textit{``wait,"} \textit{``wait for cue,"} \textit{``eye contact,"} \textit{``pause,"} 
and \textit{``speak."} The skill's starting action is \textit{``approach,"} where the robot observes the human situation while approaching closer to the person. The skill's end action is \textit{``speak,"} where the robot speaks an initial phrase to start a conversation.

The defined robot goal for this skill is \textit{``Initiate a conversation at $E$"} where $E$ is a parameter that can be passed to the skill indicating the environment the skill will be used in (e.g., at \textit{``a facility"}). No other setup or implementation is needed, and the skill can run with just the information in italic. However, each action requires some additional setup (defining the action text, reference response, action stimuli) and slight implementation as shown in \tabref{tab:engage_actions}. These actions once implemented can be reused by other skills that are based on the same skill design, allowing a reusable ecosystem.



\section{Experiment}
\label{experiment}

\begin{figure*}[t]
\centering
\includegraphics[width=1.9\columnwidth]{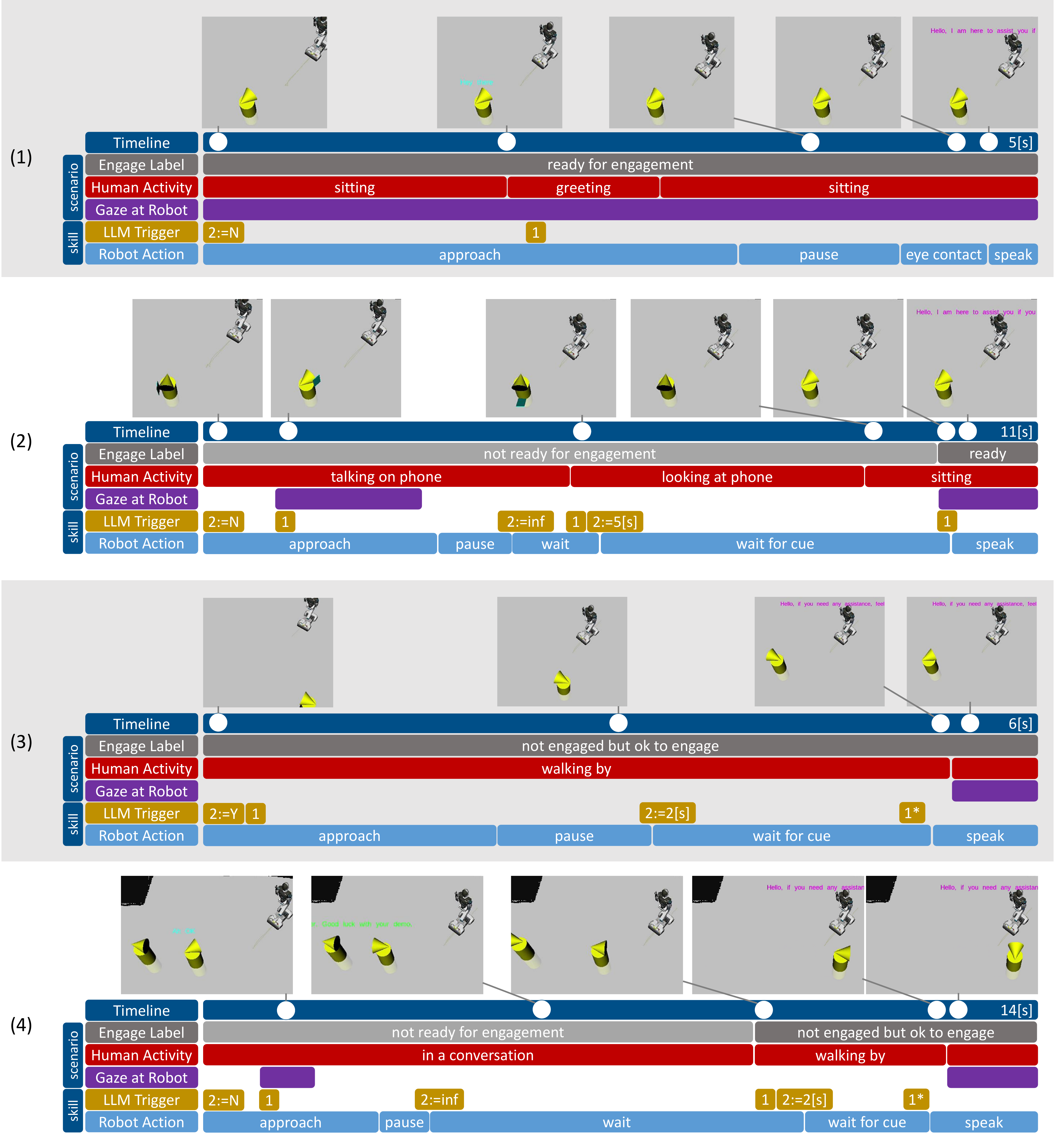}
\caption{Experiment results for the four test scenarios: (1) person-robot (a person sitting waited to talk with the robot), (2) person-object (a person talking on a phone when the robot comes), (3) person-environment (a person walking by not noticing the robot), (4) person-person (a person in a conversation then walking out). Timeline indicates start-to-end from left-to-right, with the duration of the scenario on the very right of the line. Engage Label refers to the time-sections of the ground-truth labels. Human Activity refers to the activity description published from the simulator. Gaze at Robot indicates the time-sections where a ``looking toward" description is published from the simulator (otherwise ``looking away" is published). LLM Trigger indicates the timings the LLM component is called by the event manager, where 1 indicates calling for the first-stage action decision, and 2 indicates calling for the second-stage timing decision. Y/N refers to the yes/no response for the chance of losing the person question, $T$[s] refers to the returned wait time where inf indicates an infinite wait. 1* indicates timings where the LLM was called after waiting the wait time. An action changing without the call of the LLM indicates that the previous call returned a sequence of actions (e.g., ``The robot should stop, make eye contact, and respond with a friendly phrase."). }
\label{fig:experiment}
\vspace{-3mm}
\end{figure*}

We test the skill design by applying the engage skill in four different scenarios. The situation for each scenario will differ in that, at the beginning of each scenario, the human's attention is pointed toward a different target. That is, we test the skill in the following four representative person-$X$ scenarios, where $X$ refers to the target of pointed attention: (1) person-robot, (2) person-object, (3) person-environment, (4) person-person. Each scenario contains a time-series of human actions and a time-section with ground-truth labels about the person's engagement state: ``not ready for engagement," ``not engaged but ok to engage," or ``ready for engagement." All scenarios contain at least one time section with a ``not engaged but ok to engage" or ``ready for engagement" label. The execution of the skill is considered successful if the skill runs the end action (speak) during those time sections. The execution is considered a failure if the skill runs the end action in a  ``not ready for engagement" section or does not run the end action by the time of the last human action in the scenario (i.e., a timeout).

The test is done in simulation to run multiple times under a consistent testing condition. Each scenario is ran ten times in order to evaluate the reproducibility of the LLM's output. The human activity and gaze directions do not contain any noise, and the ground truth description is published from the simulator. The description of the human activity, however, is based on actual descriptions generated using a VLM \cite{abdin2024phi}. The descriptions were obtained by capturing a video of the situation in the real world and then processing the video frames with the VLM.

The robot is controlled using a simulated controller and lidar of the simulated environment. Communication between the ground truth descriptions, event manager, and robot control uses ROS and ROS2 \cite{ros2}. The skill is built on top of the open-source task-sequencer framework \cite{tss2024} to be combined with other skills in a longer operation.

\subsection{Skill Performance}

\begin{table}[t]
\caption{Performance of the proposed skill design on four scenarios tested ten times each. Numbers indicate success over trials.}
\vspace{-3mm}
\label{tab:performance}
\begin{center}
\begin{tabular}{|l|l|}
\hline
scenario & success rate \\
\hline
person-robot & 10 / 10  \\
\hline
person-object & 10 / 10 \\
\hline
person-environment & 8 / 10 \\
\hline
person-person & 8 / 10 \\
\hline
\end{tabular}
\end{center}
\end{table}

\begin{table}[t]
\caption{Average frequency of each action in each scenario. Averaged over the ten trials.}
\vspace{-3mm}
\label{tab:frequency}
\begin{center}
\begin{tabular}{|l|c|c|c|c|}
\hline
action & -robot & -object & -environment & -person \\
\hline
approach & 1.0 & 1.0 & 1.0 & 1.0  \\
\hline
wait & 0.0 & 1.0 & 0.0 & 1.0  \\
\hline
wait for cue & 0.0 & 1.0 & 1.2 & 1.2 \\
\hline
eye contact & 0.7 & 0.3 & 0.0 & 0.0 \\
\hline
pause & 0.7 & 1.0 & 1.0 & 1.0  \\
\hline
speak & 1.0 & 1.0 & 0.8 & 0.8  \\
\hline
\end{tabular}
\end{center}
\vspace{-4mm}
\end{table}

\begin{table}[t]
\caption{Performance without the proposed action text.}
\vspace{-3mm}
\label{tab:no_text_result}
\begin{center}
\begin{tabular}{|l|l|}
\hline
scenario & success rate \\
\hline
person-robot & 10 / 10  \\
\hline
person-object & 4 / 10 \\
\hline
person-environment & 0 / 10 \\
\hline
person-person & 0 / 10 \\
\hline
\end{tabular}
\end{center}
\end{table}

\begin{table}[t]
\caption{Average frequency of each action without the proposed action text.}
\vspace{-3mm}
\label{tab:no_text_freq}
\begin{center}
\begin{tabular}{|l|c|c|c|c|}
\hline
action & -robot & -object & -environment & -person \\
\hline
approach & 1.0 & 1.6 & 1.0 & 1.0  \\
\hline
wait & 0.0 & 1.4 & 0.0 & 1.0  \\
\hline
wait for cue & 0.0 & 0.6 & 2.0 & 2.0 \\
\hline
eye contact & 0.0 & 0.0 & 0.0 & 0.0 \\
\hline
pause & 0.0 & 1.8 & 0.1 & 0.9  \\
\hline
speak & 1.0 & 0.4 & 0.0 & 0.0  \\
\hline
\end{tabular}
\end{center}
\vspace{-4mm}
\end{table}

Details of each scenario and successful execution results are shown in \figref{fig:experiment}. As shown in the figure, the skill was able to execute all scenarios successfully. We can also observe that the second-stage questions were effective especially for the person-environment and person-person scenario, where the person did not make eye contact, and the skill would have been stuck at the \textit{``wait for cue"} action without the wait time generated by the second-stage question to the LLM. The skill would have also not completed the task for the person-environment scenario, as without the second-stage question, the skill would have lost the person by the time the \textit{``approach"} action was finished.

The overall success rate of the ten trials is shown in \tabref{tab:performance}. 
As shown in the table, the skill is executed successfully for 90\% of the trials. However, there were a few trials in the person-environment and person-person scenarios that failed by not reaching the end action and timed out by entering the \textit{``wait for cue"} action twice (which totals to a four seconds wait time based on the secondary question mentioned in \secref{stimuli}). Despite being the same situation input, there were two response patterns generated by the LLM: one where the skill performed an action sequence \textit{``approach"} then \textit{``pause"} then \textit{``wait for cue"} then \textit{``speak"} and the other \textit{``approach"} then \textit{``pause"} then \textit{``wait for cue"} then \textit{``wait for cue."} This slight inconsistency in response is what led to the failure cases.

\tabref{tab:frequency} shows the average number of times an action appeared in each scenario. A number of 1.0 indicates that the action appeared once in all trials. A number lower or higher than 1.0 indicates an inconsistency in the LLM's response, where a lower number indicates that the action did not appear in some of the tirals and a higher number indicates that the action appeared multiple times in some of the trials. From the table, we can observe some inconsistency in the person-robot scenario which also had two response patterns: one where the skill performed an action sequence \textit{``approach"} then \textit{``pause"} then \textit{``eye contact"} then \textit{``speak"} and the other directly \textit{``approach"} then \textit{``speak."} While both patterns successfully run the end action, pausing the approach before speaking could be a more natural way of interacting.

Nevertheless, the skill provides a high success rate and mostly consistent outputs. It is important to note that the skill was not tuned for each scenario and the exact same skill was used in all test scenarios. The skill was able to generate different actions for each scenario based on the difference in observed situation. If such a general skill were to be developed manually, an engineer would have to code more than 120 situation-action pairs just for the four test scenarios (and increases infinitely as new human situations are observed). The results indicate that by leveraging LLMs for the action planning, one can achieve different scenarios with no extra coding at a reasonable success rate.


\subsection{Comparison to without the Action Text}

To better understand the effect of the action text explained in \secref{llm_component}, we compare with a baseline test condition where the action text in \tabref{tab:engage_actions} is not passed to the LLM component. The results of the baseline condition is shown in \tabref{tab:no_text_result} and \tabref{tab:no_text_freq}. As shown in the table, the success rate significantly drops for the person-object, person-environment, and person-person scenario. In addition, the robot's approach is never paused for the person-robot scenario.

The main reason for the failure in the person-object and person-environment scenario is that the LLM generated a consistent action of \textit{``wait for cue"} when the person's activity was \textit{``walking by."} Therefore, when there was no change in the observered human's situation, the LLM kept returning \textit{``wait for cue"} and never reached the end action. In the case of the action text, information that the robot has been waiting for a cue is passed to the LLM, which facilitates the robot to try a different action \textit{``speak."} In a sense, the action text acted as a trigger to switch from the robot being in a passive interaction state (respecting the current human's activity) to an active interaction state (prioritizing the robot's goal/task). Without this trigger, the robot will never finish its task as the LLM will keep respecting the human's activity.  

Furthermore, without the action text, due to the lack of context, actions such as \textit{``pause"} may or may not be triggered as the LLM has no idea of whether the robot is in a static position or is currently moving. This lack of context generate some redundant actions and produce inconsistent results, which is apparent with the person-object scenario in \tabref{tab:no_text_freq}. The cases where the baseline condition failed in the person-object scenario was when the LLM redundantly tried to \textit{``approach"} the person (even after pausing, which is the same as resuming the approach), thus timed out before reaching the end action. In the case of the action text, information such as \textit{``The robot is waiting for an eye contact from the person."} clarifies the state of the robot, but without the action text, the LLM has no idea of whether the robot is already trying to initiate an interaction with the human or whether the situation is before the robot even tried to initiate the interaction.


\section{Discussions and Limitations}
\label{discussions}

LLMs indeed reduce the burden of programming / designing heuristics for deciding the action appropriate for the HRI situation. We have shown a skill design to incorporate such capabilities of LLMs, where a robot deals with the HRI situation at runtime. The engage skill designed using our skill design was applicable across four distinct scenarios and obtained a 90\% success rate. Results from the experiment prove the necessity of the design choices: second-stage questions to ask about timing of the planning, and passing action text about the current robot's action. The choice of using a starting action and end action comes from the problem setting where the robot is trying to perform a task it must complete. From the experiment, it is apparent that to generalize across various HRI situations, the robot cannot just run the end action (e.g., \textit{``speak"} for the engage skill) and it is necessary to run different actions before completing the task, such as \textit{``pause"} and \textit{``wait"} based on the human situation. The sufficiency of the action set is guaranteed with the bottom-up approach, where, if an action is ever missing, the action can easily be added with the skill design. While the experiment results support the validity of the design, we further discuss its appropriateness and limitations.

First, the second-stage question about timing has been shown necessary, but a question is whether this design choice is sufficient. The second-stage questions are highly reliant on the LLM response categories introduced in this work, and the sufficiency is dependent on whether these categories are a sufficient set. Between category 1 and 2 from \secref{stimuli}, it is apparent that any response from the LLM falls in either of the two categories: if the response contains a specific timing to observe a situation then 1, otherwise 2. It is also apparent that any action in the response of category 2 falls into either 2-A or 2-B: if the action is low-level then 2-A, otherwise 2-B. Therefore, all responses fall into either one of the categories 1, 2-A, 2-B. However, there are a few limitations in order to say that this categorization is a sufficient set. Category 1 assumes that the returned actions do not have a clear action to execute (e.g., the action \textit{``wait"} does not execute a real action), thus not requiring a further categorization about actions. Category 2 separates 2-A and 2-B as these two have different characteristics regarding the length of the execution. If a response is returned beyond these assumptions (e.g., if other characteristics must be considered for defining plan timings), further categorization of the responses are required, and different questions will be needed for each additional category.

Second, there remains a huge question of whether the action text is sufficient for facilitating the robot to prioritize its task. The action text tested in this work does not explicitly tell the robot to prioritize its task over respecting the human's current activity. However, based on findings in our previous investigation \cite{sasabuchi2025agreeing}, it is possible to explicitly guide the LLM to prioritize the robot's task by adding strong words and telling that the task is urgent.
Yet, this approach would require adding additional rules when generating action texts, such as rules of deciding when to add the strong words. If the strong words are added from the beginning, the robot will never respect the human's activity, which is not what we would expect for a skill that generalizes across different HRI situations.

Another way to think about the LLM's response is that, if the robot does not prioritize its task even with the action text, then from the LLM's common sense perspective, the robot should not prioritize its task in that situation. The current skill design and the tested scenarios assume that the robot should reach the end action, but if the situation was not that important (e.g., in a situation where a robot comes across a person the robot knows and the robot wanted to say ``hi" but the person looked extremely busy), never prioritizing the robot's task could indeed be the correct answer. Thus, the design might already be sufficient to achieve what the skill needs to achieve. This implies some design challenges for skills using LLMs. There is always the decision of how much to rely on the current LLM's responses versus how much to further guide the LLM's response. The current skill design is designed to sufficiently guide task prioritization for the cases shown in this work, but one may find this to be too much guiding or too less when applying to other cases.

A third question is whether a bottom-up approach for creating an action set is necessary. One may argue that the actions for the engage skill (instead of discovering bottom-up) could be pre-defined as is mostly about waiting, gesturing, or related to the control of an action (e.g., pausing to control the pausing of the approach action). Pre-defining the action set could simplify the problem by having the LLM choose a sequence of actions from a defined set of actions (instead of generating one). If an engineer could come up with all valid actions, then the bottom-up approach is not necessary, however, there is no guarantee that an engineer will come up with all actions required for a task involving HRI. The engage skill is only one example of applying the design and there could be more action patterns when applying the design to another HRI-involved skill. The bottom-up approach allows finding \textit{``a better suggestion"} when no valid actions in the existing action set applies to the current situation. Suggestions are one of the major benefits of using artificial intelligence (AI) in the current era and an advantageous way of leveraging AI.



\section{CONCLUSIONS}

In this work, we have shown a plan-and-act skill design leveraging LLMs to generate action plans for tasks involving situational HRI. The design enables leveraging LLM outputs and integrate bottom-up action sets, call LLMs at an appropriate timing during a runtime execution, and enables the LLM to balance between ``respecting the human activity" and ``prioritizing the robot's task goal." Results from the experiment indicate that an action text is critical for generating consistent actions, aligning with the situational context, but more importantly, facilitating the robot to accomplish its 
task in an appropriate manner when the human situation is not aligned with the robot's task goal. By leveraging LLMs, the skill can easily scale to different human situations and different HRI scenarios with a reasonable success rate. While the results prove the necessity of the design, the sufficiency of the design depends on the degree one wishes to rely on the LLM's response. Relying versus guiding responses is a design challenge for skills integrating LLMs.

As for future directions, there remains a question on how well the skill design can be applied to other HRI involved skills. This could include turn-taking during a conversation skill or a delivering skill where the robot must decide among different actions depending on the situation of the human (such as handover if the person is stretching their hands or place the delivery on a desk if the person is in a phone call). So far, the proposed skill design uses LLMs for planning and figuring out the timing of actions. The design has not focused on using LLMs for figuring out ``how to run the actions" (e.g., such as generating trajectories). Past HRI research have focused on trajectory generation by taking into account the human situation \cite{sisbot2005navigation, strabala2013toward}. An interesting question is whether LLMs can automate such implementations at the level of trajectories based on their large data of common sense. 










\end{document}